\documentclass[conference, letter]{IEEEtran}

\usepackage{graphicx}           
\usepackage{amsmath,amsfonts}   
\usepackage{multirow}           
\usepackage{algorithmic}       
\usepackage{textcomp}           

\usepackage{booktabs}           
\usepackage{adjustbox}          
\usepackage{float}              

\usepackage[table]{xcolor}      
\definecolor{lightgreen}{rgb}{0.85, 1.0, 0.85}

\usepackage{caption}            
\usepackage{subcaption}         

\usepackage{tcolorbox}          
\usepackage{hyperref}

\usepackage[linesnumbered,algoruled,boxed,lined]{algorithm2e}

\usepackage{subcaption}
\usepackage[utf8]{inputenc}
\usepackage{flushend}
\SetKwRepeat{Do}{do}{while}
\usepackage{orcidlink}
\let\oldnl\nl
\newcommand{\nonl}{\renewcommand{\nl}{\let\nl\oldnl}}
\usepackage{float}
\newtcolorbox{custombox}[1]{
	colback=gray!20,
	colframe=gray!50,
	left=1mm,
	right=1mm,
	top=1mm,
	bottom=1mm,
	fonttitle=\bfseries,
	arc=2mm,
	leftrule=0mm,
	rightrule=.5mm,
	toprule=0mm,
	bottomrule=.5mm,
	notitle,
	before=\par\smallskip\noindent,
	before upper={\textbf{#1: } },
}

\usepackage{amssymb}
\newsavebox\CBox

\newcommand{\chunk}[2]{%
\fcolorbox{black}{yellow}{\bfseries\sffamily\scriptsize#1}%
{$\blacktriangleright$#2$\blacktriangleleft$}%
}

\usepackage{listings}
\usepackage{xcolor}

\lstdefinestyle{oracleStyle}{
    language=Lisp,
    basicstyle=\ttfamily\footnotesize,
    numbers=left,                      
    numberstyle=\tiny\color{gray},     
    numbersep=8pt,                     
    backgroundcolor=\color{white},
    frame=single,                      
    rulecolor=\color{black},
    frameround=ffff,                   
    keywordstyle=\color{blue!70!black},
    commentstyle=\color{gray!70},
    stringstyle=\color{green!50!black},
    breaklines=true,
    breakatwhitespace=false,
    keepspaces=true,
    showspaces=false,
    showstringspaces=false,
    showtabs=false,
    tabsize=2,
    captionpos=b,
    abovecaptionskip=6pt,
    belowcaptionskip=6pt,
    xleftmargin=2em,                   
    framexleftmargin=1.5em
}
\newcounter{finding}
\newcommand\pimodel{$\pi_0$} 
\newtcolorbox{FindingBox}[1][]{%
  colframe=black!50,
  colback=white,
  sharp corners,
  left=4pt,
  right=4pt,
  top=2pt,
  bottom=2pt,
  boxsep=2pt,
  before skip=6pt,
  after skip=6pt,
  fontupper=\small,
  before upper={%
    \stepcounter{finding}%
    \textbf{Finding~\thefinding. }%
  },
  #1
}

\newcommand{\aitor}[1]{\chunk{Aitor}{\textbf{\textcolor{red}{\textsl{#1}}}}}
\newcommand{\pablo}[1]{\chunk{Pablo}{\textbf{\textcolor{blue}{\textsl{#1}}}}}

\begin{document}

\title{Metamorphic Testing of Vision--Language Action–Enabled Robots}

\author{\IEEEauthorblockN{Pablo Valle}
\IEEEauthorblockA{
\textit{Mondragon University}\\
Mondragon, Spain\\
pvalle@mondragon.edu}
\and
\IEEEauthorblockN{Sergio Segura}
\IEEEauthorblockA{
\textit{University of Seville}\\
Seville, Spain\\
sergiosegura@us.es}
\and
\IEEEauthorblockN{Shaukat Ali}
\IEEEauthorblockA{
\textit{Simula Research Laboratory}\\
Oslo, Norway\\
shaukat@simula.no}
\and
\IEEEauthorblockN{Aitor Arrieta}
\IEEEauthorblockA{
\textit{Mondragon University}\\
Mondragon, Spain\\
aarrieta@mondragon.edu}}

\maketitle

\begin{abstract}

Vision-Language-Action (VLA) models are multimodal robotic task controllers that, given an instruction and visual inputs, produce a sequence of low-level control actions (or motor commands) enabling a robot to execute the requested task in the physical environment. These systems face the test oracle problem from multiple perspectives. On the one hand, a test oracle must be defined for each instruction prompt, which is a complex and non-generalizable approach. On the other hand, current state-of-the-art oracles typically capture symbolic representations of the world (e.g., robot and object states), enabling the correctness evaluation of a task, but fail to assess other critical aspects, such as the quality with which VLA-enabled robots perform a task. In this paper, we explore whether Metamorphic Testing (MT) can alleviate the test oracle problem in this context. To do so, we propose two metamorphic relation patterns and five metamorphic relations to assess whether changes to the test inputs impact the original trajectory of the VLA-enabled robots. 
An empirical study involving five VLA models, two simulated robots, and four robotic tasks shows that MT can effectively alleviate the test oracle problem by automatically detecting diverse types of failures, including, but not limited to, uncompleted tasks. More importantly, the proposed MRs are generalizable, making the proposed approach applicable across different VLA models, robots, and tasks, even in the absence of test oracles.

\end{abstract}


\section{Introduction}
Vision-Language-Action (VLA) models are a new family of multimodal control techniques that, given a natural-language instruction and a set of images, convert high-level semantic commands into low-level control actions for a robotic system~\cite{zitkovich2023rt, openvla:online, nvidia2025gr00tn1openfoundation}. These new types of Artificial Intelligence (AI) techniques are expected to become the next generation of control algorithms, enabling generalist robots (e.g., humanoids) to flexibly perform a wide variety of tasks in unstructured environments. Different VLA models and architectures exist~\cite{ma2024survey}, and the robotics and AI communities are making significant efforts to ensure that these models are both efficient, enabling inference at the edge to control physical robotic arms in real-time, and effective for task execution.

\begin{figure*}[ht]
    \centering
    \includegraphics[width=\linewidth]{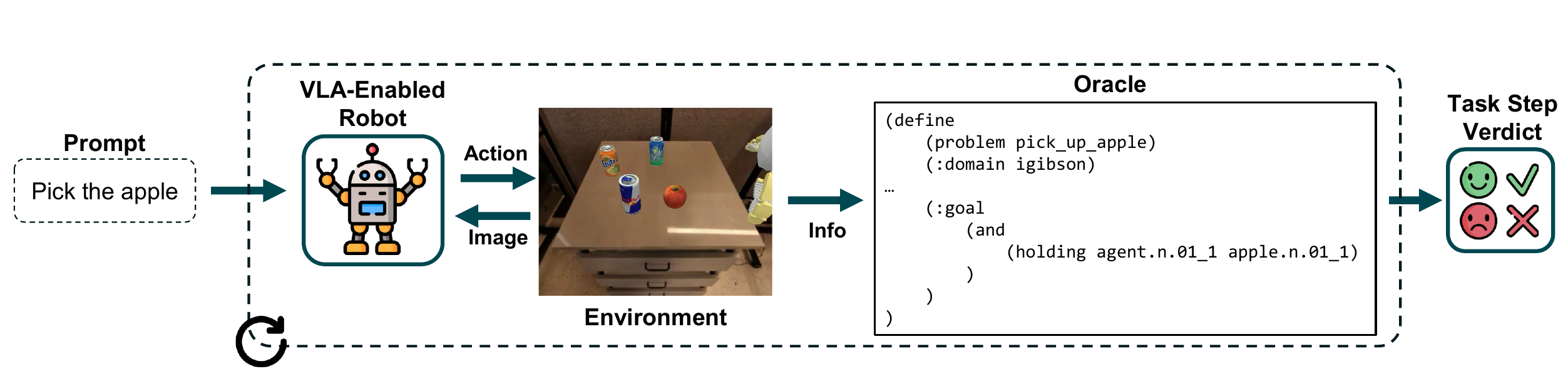}
    \caption{Task execution pipeline for the ``\textit{pick the apple}'' task}
    \label{fig:oracle}
\end{figure*}

Despite these recent advances, the evaluation of VLA models largely depends on benchmarks that employ symbolic oracles~\cite{khazatsky2024droid, li24simpler}, i.e., oracles that, given an instruction in natural language, check whether the symbolic states of the objects and the robotic system satisfy the instruction. Figure~\ref{fig:oracle} depicts an example of symbolic oracles in the test execution pipeline for the ``\textit{pick the apple}'' instruction. The control process begins with a natural-language instruction prompt provided to the VLA model that controls the robot. At each execution step, the VLA model processes a visual observation of the environment to generate an action command, which the robot executes. An oracle, with access to information about the objects in the environment, subsequently evaluates whether the task has been completed successfully. The symbolic oracle encodes the expected goal state in a symbolic formalism. For instance, for the ``\textit{pick the apple}'' prompt, the symbolic oracle specifies that the agent should end the episode holding the target object. These oracles suffer from two core problems. First, the automated generation of symbolic test oracles depends heavily on the task instruction. Current techniques (e.g., VLATest~\cite{wang2025vlatest}) rely on predefined task and object templates to automatically generate these kind of oracles. However, the way humans instruct robots is highly variable, and the same task can be expressed in multiple ways (e.g., \textit{``pick up the apple''},\textit{``grab the apple''}, or \textit{``take the apple from the table''}). While Large Language Models (LLMs) could be potentially used to generate test oracles automatically, they are prone to hallucinations, and human instructions may be inherently ambiguous or underspecified. Second, symbolic oracles fail to capture qualitative aspects of task execution~\cite{valle2025evaluating}. For instance, they do not assess whether the trajectory generated from the actions provided by the VLA model is optimal, nor whether object manipulation is performed correctly and safely (e.g., whether objects are dropped or whether there are any kind of collisions in the scene). As a result, important classes of failures related to motion quality, stability, or suboptimal behavior remain undetected.

\vspace{0.1cm}
In practice, previous limitations mean VLA models suffer the oracle problem, that is, assessing whether their output is correct beyond trivial and domain-specific cases is hardly feasible~\cite{2014-barr-tse}. Metamorphic Testing (MT) addresses the oracle problem by shifting the focus from validating individual program executions to verifying whether expected relations, known as metamorphic relations (MRs), hold across multiple executions under related inputs. Instead of requiring explicit output correctness criteria, MT leverages properties that should be satisfied when inputs are systematically varied, such as expecting similar behaviour from a self-driving car when traversing the same route under different non-extreme weather conditions~\cite{tian18-icse}. To further support the identification of MRs, researchers have introduced metamorphic relation patterns (MRPs), which abstractly characterize families of related MRs and guide testers in systematically deriving specific relations~\cite{2018-zhou-tse,2018-segura-met,2017-segura-tse}. 

In this paper, we investigate whether MT can alleviate the test oracle problem in this domain. To this end, we define two MRPs that rely on the robotic trajectories produced by VLA models. These MRPs enable the derivation of MRs for VLA-enabled robots and the automated execution of test cases without symbolic oracles. We define five MRs and evaluate them across five different VLA models, two robots, and four representative tasks. We analyze and compare the effectiveness of MT and each of the proposed MRs in 9,320 source-follow-up test case pairs. Furthermore, we analyze the differences in the failures identified by the proposed MRs and symbolic oracles, and we propose a taxonomy of failures for VLAs. Our evaluation suggests that effective testing of VLA-enabled robotic systems requires combining symbolic oracles with metamorphic testing under carefully selected strictness thresholds and multiple metamorphic relations to capture both task-level correctness and execution-level robustness. In addition, we provide a complete replication package~\cite{MT_replication_package}, including code, configuration files, and instructions to facilitate reproducibility and further research.

The rest of the paper is structured as follows: in Section~\ref{sec:Back_VLA}, generic background of VLAs and MT are provided. Section~\ref{sec:MRIPs} proposes our MRPs and the derived five MRs. Section~\ref{sec:empirical_study} explains the empirical study we conducted, followed by Section~\ref{sec:results}, which reports and discusses the results and Section~\ref{sec:threats}, which explains how we mitigated the threats. In Section~\ref{sec:relatedwork} we position our work with respect to the state-of-the-art and we conclude the paper in Section~\ref{sec:conclusion}.

\section{Background}
\label{sec:Back_VLA}
In this section we present basic background related to VLA models followed by general theory of MT.
\subsection{Vision-Language-Action Models}

Vision-Language-Action (VLA) models~\cite{nvidia2025gr00tn1openfoundation,black2024pi0visionlanguageactionflowmodel,kim2024openvla,qu2025spatialvla, qu2025eo1} represent an emerging class of multimodal learning systems that integrate perception, language understanding, and trajectory planning. Unlike traditional deep neural networks (DNNs), which are typically specialized for single-domain tasks, such as Convolutional Neural Networks (CNN) which excel at extracting hierarchical features from images \cite{lecun2002gradient,krizhevsky2012imagenet,simonyan2014very,szegedy2015going}, and transformer-based Large Language Models (LLMs) which capture long-range dependencies in text \cite{vaswani2017attention,achiam2023gpt,Claude:online,team2023gemini,grattafiori2024llama,jiang2023mistral7b}, VLA models encode visual observations, such as images or video frames, and textual instructions into a shared embedding space. From this shared representation, they generate structured action sequences that form a coherent trajectory, enabling them to interpret commands like ``\textit{Pick the apple}’’ within a scene. This requires resolving ambiguities in object references through visual attention mechanisms and translating high-level instructions into precise physical movements.

To formalize this process, a VLA generates a sequence of action chunks ($AC_t$) that, when executed on the robot, produce a trajectory ($\mathcal{T}$) accomplishing a predefined task. Let $A = \left[ AC_{t},\, AC_{t+H},\, \dots,\, AC_{t+(N-1)*H} \right]$ denote the sequence of action chunks produced by the VLA, where each $AC_t$ is a temporally extended control command issued at time step $t$, $N$ is the number of action chunks required to complete the task and $H$ is the time-step horizon of each action chunk. At each decision step $t$
the VLA provides one action chunk, a sequence of future robotic actions conditioned on the current observation ($o_t$), expressed as $p(AC_t \mid o_t)$. In other words, the model translates the integrated information from instructions and visual input into actionable motor commands. An action chunk $AC_t$ is modeled as a sequence of future actions conditioned on the current observation, i.e., $AC_t = \left[a_t,\, a_{t+1},\, \dots,\, a_{t+H-1}\right]$, $H$ being the length of the action chunk (e.g., $H=50$~\cite{black2024pi0visionlanguageactionflowmodel}, $H=4$~\cite{nvidia2025gr00tn1openfoundation}). The input to the model is an observation at time step $t$, i.e., $o_t = \left[I^t_1,\, I^t_2,\, \dots,\, I^t_n,\, \ell_t,\, \theta_t\right]$, where each $I^t_i$ is an RGB image at time $t$, $\ell_t$ represents a language instruction, and $\theta_t$ denotes the robot’s proprioceptive state. These inputs are first encoded and then projected into a shared embedding space. When the robot executes these actions chunks, they induce a trajectory $\mathcal{T} = \left[ P_t,\, P_{t+1},\, \dots,\, P_{t+T} \right]$, such that each $P_t$ represents the robot's end effector's state at time step $t$, including its position ($x,y,z$) and orientation in ($q_1, q_2, q_3, q_4$). This way, VLAs serve as the core mechanism for controlling the robot, integrating the provided instruction with visual information from its environment to generate the action sequences.

\begin{figure}[h]
    \centering
    \includegraphics[width=\linewidth]{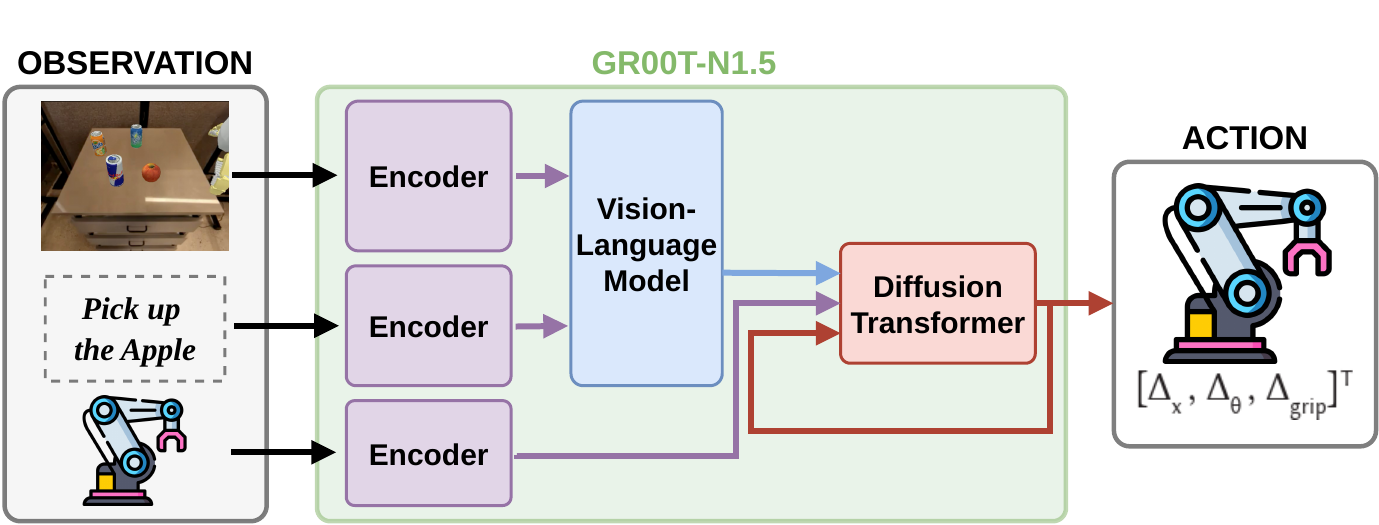}
    \caption{Architecture of GR00T-N1.5~\cite{nvidia2025gr00tn1openfoundation}}
    \label{fig:arch}
\end{figure}

Recent VLA models~\cite{nvidia2025gr00tn1openfoundation,black2024pi0visionlanguageactionflowmodel} incorporate a diffusion-based denoising mechanism, implemented through a diffusion transformer before producing the action chunk, as depicted by Figure~\ref{fig:arch}. In these models, the action chunk $A_t$ generated by the VLA is progressively refined via an iterative denoising diffusion process. At each denoising step, the diffusion transformer leverages contextual representations of the visual observation, language instruction and propioceptive state, allowing the model to correct and adjust the initial action prediction. Rather than generating actions from scratch, the diffusion module operates as a structured post-processing component that steers the predicted actions towards more consistent regions of the action space. This architecture enables the model to better handle complex scenes and increase robustness by reducing the error accumulation in long-horizon tasks ~\cite{black2024pi0visionlanguageactionflowmodel,zhao2023learning, nvidia2025gr00tn1openfoundation}.

\subsection{Metamorphic Testing}

\emph{Metamorphic testing} (MT)~\cite{1998-chen-tr,2016-segura-tse,2017-chen-cs} is a testing approach aimed at revealing defects by analysing relationships between the inputs and outputs of multiple executions of the program under test. These relationships are known as \emph{metamorphic relations} (MRs). MT is particularly useful in the presence of the oracle problem, where determining the correctness of outputs for complex inputs is difficult or impractical. As an example, consider a program $detect(I)$ that searches for clinically relevant patterns, such as lesions or anomalies, in a medical image $I$. For realistic medical images, directly assessing the correctness and completeness of the detected patterns may be infeasible, which constitutes an instance of the oracle problem. Suppose a new image $I'$ is constructed by augmenting the original image $I$ with an additional, independent image region $R$, such that $I' = I + {R}$. Intuitively, all patterns detected in $I$ should also be detected in $I'$. This expectation can be formalised as the MR $detect(I) \subseteq detect(I')$, where $I$ is the source test case and $I'$, obtained by extending the input image, is the corresponding follow-up test case. This relation can be instantiated into one or more \emph{metamorphic tests} by selecting specific images and verifying whether the relation holds. A violation of the relation indicates a failed metamorphic test and suggests the presence of a defect in the program. MT has been successfully applied across a wide range of domains, including web services, compilers, cybersecurity, autonomous vehicles, and bioinformatics~\cite{2016-segura-tse,2017-chen-cs}, with reported industrial adoption at Adobe\cite{2017-jarman-met}, Google~\cite{donaldson19-met}, and Meta~\cite{ahlgren21-icse}.

MRs are often specified at a high level of abstraction, describing not a single relation but a family of related MRs. Such abstractions are referred to as \emph{metamorphic relation patterns} (MRPs)~\cite{2018-zhou-tse,2018-segura-met,2017-segura-tse}. Zhou et al.~\cite{2018-zhou-tse} define a MRP as an abstraction that characterizes a set of (possibly infinitely many) MRs and demonstrate their usefulness in supporting testers during the identification of suitable relations. One example is the symmetry MRP, motivated by the observation that many systems exhibit invariant behaviour when observed from different perspectives. For instance, an AI-based object recognition system should detect the same objects in a video regardless of whether the video is played forwards or backwards. Similarly, Segura et al.~\cite{segura19-met} introduced several MRPs for query-based systems, such as adding restrictive filters to a query and expecting the resulting output to be a subset of the original results.

MRPs can be further divied into two subtypes: \emph{metamorphic relation input patterns} (MRIPs)~\cite{2018-zhou-tse} and \emph{metamorphic relation output patterns} (MROPs)~\cite{2017-segura-tse}. MRIPs define only the relationships among inputs, leaving the relationships among outputs unspecified. Conversely, MROPs specify the relationships among outputs while leaving the relationships among inputs undefined. For example, 
Segura et al.~\cite{2017-segura-tse} introduced the subset MROP, which characterises relations in which the output of a follow-up test case is expected to be a subset of the source output. This can be achieved by changing inputs in multiples ways, for example, by adding filters or restricting the search strings in a search engine. In this paper, we present two MRPs together with five MRs derived from them for VLA-enabled robots.

\section{Metamorphic Relation Patterns}\label{sec:MRIPs}

In this section, we propose two novel MRPs: Trajectory Consistency (TC) and Trajectory Variation (TV). These patterns represent groups of MRs that involve changes to the source input, which either should not affect the trajectory of the robot (i.e., TC) or have a predictable change on the trajectory of the robot (i.e., TV) during test case execution. 

The intuitive idea behind these patterns is that, in VLA-enabled robots, it is often straightforward to anticipate how changes in certain inputs impact the trajectory. For instance, relocating a target object in the workspace should result in a motion path that adapts to reach the new position, assuming that the VLA correctly interprets spatial relations. Conversely, adding an irrelevant object that does not intersect the planned path should not significantly alter the trajectory of the robot. If the trajectory fails to change when it should, or changes when it should not, this signals a potential failure in the VLA perception, planning, or interpretation of language actions.

A key characteristic of trajectory-based MRPs is their generalization, which makes them potentially applicable to a wide range of robots and tasks. Unlike symbolic oracles, which provide a binary assessment of the task success (i.e., whether the task has been performed or not), MRPs assess how the robot behaves throughout the task. This enables the assessment of behavioral consistency, safety and robustness over time, rather than focusing only on final outcomes. This makes symbolic oracles and MT complementary: while symbolic oracles capture task-level correctness, MT captures behavioral properties. From the software testing perspective, MT alleviates the oracle problem, which is particularly pronounced in this domain where specifying complete and unambiguous correctness criteria is cumbersome. Symbolic oracles are task-specific and often ambiguous in robotic domains, whereas the MRs proposed in this paper are task-agnostic and easier to define, as they encode expected relationships between executions rather than absolute correctness conditions which enables them to be reused and scalable between different test designs. As a result, MR-based evaluation can reveal misbehaviors that remain invisible under binary evaluation alone. MT assesses whether the relation between the source and the follow-up trajectories matches the expected behavior specified by the MR. Depending on the MR, the follow-up trajectory may need to adapt predictably or remain invariant relative to the source trajectory. 



\subsection{MRP1: Trajectory Consistency Pattern (TC)}

The TC captures input transformations that should not affect the robot's trajectory. This pattern is designed to evaluate the model's robustness to superficial changes in instructions or environmental conditions, ensuring that the robot maintains predictable and stable behavior when task semantics remain unchanged. We derived three different MRs from this pattern, namely:

\textbf{MR1: Synonym Substitution.} In this MR, the action verb in the prompt is replaced with a synonym while preserving its meaning (e.g., ``\textit{Pick the apple}'' $\rightarrow$ ``\textit{Grab the apple}''). Therefore, this change should not affect the robot's trajectory 
Let $P_s$ and $P_f$ denote the source and follow-up prompts, and $\tau(T, E)$ the trajectory produced by the robot, so this MR can be represented as follows:
\begin{equation}
    d_F(\tau(P_s, E), \tau(P_f, E)) \lesssim \delta
\end{equation}

where $d_F$ is the distance between the trajectories of the robot for the source and follow-up test cases. In this work, we use the Fréchet distance, although other distance measures are also applicable (see Section~\ref{sec:empirical_study} for details). In practice, non-determinism may lead to differences, with $\delta$ representing the accepted tolerance level for an MR to be considered satisfied or, otherwise, violated.\noindent Violations of this MR represent sensitivity to lexical variations that are irrelevant to task executions, highlighting limitations in the semantic grounding of the VLA.

\textbf{MR2: Non-Interfering Object Addition.} Adding objects to the environment far from the target object should not alter the robot's trajectory. Let $E_s$ be the source environment configuration and $E_f$ the follow-up environment configuration, including additional objects. The robot's trajectories are $\tau(P, E_s)$ for the source test case and $\tau(P, E_f)$ for the follow-up test case. This MR can be formulated as follows:

\begin{equation}
    d_F(\tau(P, E_s), \tau(P, E_f)) \lesssim \delta
\end{equation}

\noindent Violations of this MR indicate that the robot overreacts to irrelevant environment objects, revealing weaknesses in how it plans and adapts its movements. Similar MRs have been leveraged in the context of self-driving cars~\cite{zhou19-cacm}.

\textbf{MR3: Light Brightness Change.}
This MR captures input transformations affecting the VLA's visual input by altering scene illumination, without affecting the robot's performance. Formally, let $E_s$ be the source environment and $E_f$ the follow-up environment obtained by applying a small illumination transformation to $E_s$. For the same prompt $P$, the source and follow-up trajectories are $\tau(P, E_s)$ and $\tau(P, E_f)$. This MR can be stated as follows:

\begin{equation}
    d_F(\tau(P, E_s), \tau(P, E_f)) \lesssim \delta
\end{equation}

\noindent Violations of this MR indicate that the VLA is sensitive to superficial illumination changes, revealing a lack of robust visual grounding. This type of MR have been successfully used in critical domains such as self-driving cars~\cite{tian18-icse} and road object detection systems~\cite{wozniak23-aitest}.

\subsection{MRP2: Trajectory Variation Pattern (TV)}
The TV captures input transformations that are expected to induce predictable changes in the robot's trajectory. These transformations reflect meaningful modifications in the instruction or environment configuration, and evaluating them provides insight into the robot's adaptability and reasoning.

\textbf{MR4: Negation or Task Inversion.}
In this MR, the prompt is modified to negate the original task  (e.g., ``\textit{Pick the apple}'' $\rightarrow$ ``\textit{Don't pick the apple}''), so the robot is expected to alter its trajectory or even refuse to move, i.e., the robot remains static in the same position. Therefore, this MR can be represented as follows:

\begin{equation}
d_F\big(\tau(P_s, E), \tau(P_f, E)\big) \gtrsim \delta
\end{equation}
where $\delta$ is a minimum threshold representing a meaningful trajectory change. A violation of this MR reveals weaknesses in semantic understanding of the asked task. Note that when the robot remains stationary, the distance to the trajectory is computed in the same manner as for trajectories involving motion.

\textbf{MR5: Target Object Relocation.} 
When the target object is moved in the workspace, the robot should generate a new trajectory reaching the new location. Let $p_s = (x_s, y_s)$ and $p_f= (x_s+\Delta x, y_s+\Delta y)$ be the source and follow-up positions of the target object, and let $\tau(P, E(p_s))$ and $\tau(P, E(p_f))$ be the corresponding trajectories, this MR can be represented as follows:

\begin{equation}
    \alpha \Vert \Delta p \Vert \lesssim d_F\big(\tau(P ,E(p_s)), \tau(P, E(p_f))\big) \lesssim \beta \Vert \Delta p \Vert
\end{equation}

This expresses that the translation $\Delta p$ of the target object should produce a corresponding change in the robot's trajectory. The magnitude of this translation is given by $\Vert \Delta p \Vert = \sqrt{(\Delta x)^2+(\Delta y)^2}$. However, because the robot’s kinematics and reachability constraints shape how motion plans adapt, the resulting trajectory change may not exactly match the object translation. To address this challenge, this MR introduces lower and upper proportionality bounds $\alpha$ and $\beta$, which specify the expected range of distance between the source and follow-up trajectories. The lower bound $\alpha$ ensures that the robot adapts its trajectory sufficiently in response to the object’s movement, preventing cases in which the planner fails to react meaningfully. Conversely, the upper bound $\beta$ limits the magnitude of trajectory deviation, ensuring that the robot does not overreact to the translation of the target object with unnecessary or erratic motion.

\section{Empirical Study}
\label{sec:empirical_study}

We conducted an empirical study to assess the effectiveness of MT in evaluating the behavior of VLA-enabled robots. This section presents our  research questions and details the experimental setup used for the evaluation.

\subsection{Research Questions}
In our evaluation, we aimed to answer the following Research Questions (RQs):
\begin{itemize}
    \item \textbf{RQ1 -- \textit{How effective is MT at detecting failures in VLA-enabled robots?}} With this RQ, we aimed at assessing the effectiveness of MT in detecting failures in VLA-enabled robotic systems. In particular, we compared three failure-detection strictness thresholds (i.e., low, medium, and high) based on different trajectory distances and analyzed how these thresholds influence the identification of failures. We further compared the resulting failure detection capabilities against those obtained using traditional symbolic oracles.

    \item \textbf{RQ2 -- \textit{How do different MRs differ in their ability to detect failures?}}  This RQ assesses the effectiveness of the proposed MRs in detecting failures. By comparing failure detection rates across MRs, models, tasks, and thresholds, we analyzed which types of input transformations are the most effective at detecting failures in VLA-enabled robots.

    \item \textbf{RQ3 -- \textit{What types of failures does MT reveal?}} With this RQ, we aimed at analyzing the type of failures detected by different oracle types, i.e., MT and symbolic oracles. To do so, we characterized and developed a taxonomy of failures by surveying two domain experts. We later compared the nature of the failures revealed by each oracle types and investigated whether they expose different failure types.

    

\end{itemize}

\subsection{Subject VLA Models}\label{sec:models}
For our evaluation, we selected five state-of-the-art VLA models: OpenVLA~\cite{kim2024openvla}, \pimodel~\cite{black2024pi0visionlanguageactionflowmodel}, SpatialVLA~\cite{qu2025spatialvla},  GR00T-N1.5~\cite{nvidia2025gr00tn1openfoundation}, and EO-1~\cite{qu2025eo1}. These models were chosen based on their prior usage in related studies~\cite{valle2025evaluating, wang2025vlatest, peng2025nebula, zhang2025vlabench} and because all of them are compatible with the open-source simulation environment SimplerEnv~\cite{li24simpler}. We conducted our experiments using the environment adopted from state-of-teh-art works~\cite{valle2025evaluating, wang2025vlatest}, employing the fine-tuned versions of each model adapted to two benchmark datasets corresponding to four task categories in our evaluation suite. Specifically, models for the \textit{Pick up} and \textit{Move Near} tasks were trained for the Google Robot, while models for the \textit{Put on} and \textit{Put in} tasks were trained for the WidowX robot. We evaluated the VLA models using the official checkpoints released by their respective authors. For OpenVLA, we used the base checkpoint provided on HuggingFace~\cite{openvla:online}. For \pimodel, we employed the dataset-specific checkpoints~\cite{HaomingFractal:online,HaomingBridge:online} released by the SpatialVLA authors. SpatialVLA was evaluated using the pretrained checkpoint trained on the combined datasets~\cite{IPECCOMM11:online}. For GR00T-N1.5 and EO-1, we used the fine-tuned checkpoints provided by the authors, with separate versions trained on the Fractal and Bridge datasets~\cite{gr00t_Fractal:online,gr00t_Brdige:online, eo1_fractal:online, eo1_bridge:online}.

\subsection{Environments}
Following prior work \cite{valle2025evaluating, wang2025vlatest}, we evaluate on the SimplerEnv benchmark \cite{li24simpler} across two robotic platforms and four standard manipulation tasks. The Google Robot, also known as Everyday robot\footnote{\url{https://everydayrobots.ai/}}, uses VLA models trained on the Fractal dataset \cite{brohan2022rt} and is evaluated on Pick Up and Move Near, while the WidowX robot uses models trained on the Bridge V2 dataset \cite{walke2023bridgedata} and is evaluated on Put On and Put In. Regarding the task completion constraints, the Pick Up task requires grasping and lifting the target object by at least 0.02 m for five consecutive frames; Move Near requires positioning object A within 0.05 m of object B; Put On requires stably stacking object A on object B; and Put In requires fully placing object A inside object B.

\subsection{Configurations}\label{sec:config}
To ensure reproducibility and facilitate fair comparisons with future works, we report all relevant parameter settings used in our experiments. A critical factor for replicating our results is the set of VLA model weights, described in Section~\ref{sec:models}. In addition to providing the model checkpoints, we used a fixed random seed across all the MRs and all the models to initialize the environments, ensuring that the generation of the follow-up test cases and the initial state of the environment remained consistent across all evaluations. In addition, for MR2, we set to 0.1m the distance between the non-interfering object and the target object.


We compared source and follow-up trajectories using the Fréchet distance~\cite{frechet1906quelques}. Violations were raised when distances exceeded tolerance thresholds, as reflected in the definition of the MRs~\cite{ayerdi2022performance,murphy2009automatic, 2017-segura-icsenier}. In line with related work~\cite{ayerdi2022performance}, threshold values were calibrated based on a preliminary analysis of MT results. Specifically, we defined percentile-based distance thresholds yielding three tolerance levels: strict (i.e., 0.1m, around 20th percentile) to detect subtle deviations, medium (i.e., 0.2m median) for typical variability, and low (i.e., 0.3m, around 80th percentile) to account for pronounced anomalies. For example, this means that a distance difference of 0.15m between the source and follow-up test cases in MR2 (before and after adding a non-interfering object) would be considered a violation under the strict tolerance level, but would be regarded as a satisfaction under the medium and low tolerance levels.

Regarding the selection of source test cases, we followed the approach reported by Valle et al.~\cite{valle2025evaluating}. Specifically, we selected only those source cases deemed successful by the symbolic oracle. Starting from failing test cases would have led to a biased evaluation, potentially inflating MT failure-detection results. More importantly, this approach allows us to evaluate not only the failure-detection capability of each method, but also how the two approaches complement one another, with MT leveraging test cases for which symbolic oracles fail to identify failures. This procedure yielded 116 source test cases for OpenVLA, 386 for \pimodel, 406 for SpatialVLA, 228 for GR00T-N1.5 and 728 for EO1. Based on these, we generated 580 follow-up test cases for OpenVLA, 1930 for \pimodel, 2030 for SpatialVLA, 1140 for GR00T-N1.5, and 3,640 for EO-1, resulting in a total of 1,864 source test cases and 9,320 follow-up test cases across all models. In all these follow-up test cases the symbolic oracles were also executed.

\subsection{Execution Platform and Runs}
To accelerate the execution of our experiments, we distributed the workload across two computing platforms. Experiments involving the \pimodel{} and OpenVLA models were executed on a server equipped with an AMD EPYC 7773X CPU and an NVIDIA RTX A6000 GPU with 48 GB of memory. In contrast, experiments for GR00T-N1.5 and SpatialVLA were performed on a workstation with an AMD Ryzen 9 7900X CPU and an NVIDIA GeForce RTX 4080 SUPER GPU with 16 GB of memory. Both platforms used 64-bit Ubuntu 20.04 LTS, Python 3.10, and CUDA 12.8.

Due to differences in hardware platforms as well as in the internal architectures and learned weights of the models, each VLA exhibited different levels of resource consumption, which in turn affected the execution time of each test. On average, a single test execution required approximately 26 seconds for OpenVLA, 22 seconds for \pimodel{}, 35 seconds for SpatialVLA, 7.5 seconds for GR00T-N1.5 and 12.5 seconds for EO-1. The total execution time of our experiments took 219,168 seconds, approximately 61 hours of GPU. 


\subsection{Failure characterization and taxonomy development process}

In our experiment, our oracles detected a total of 7,899 failures, making it unfeasible to manually analyze each of them. We therefore resorted to Cochran’s sample size statistical test, which provided us a statistically representative subset of failures to analyze with a predefined confidence level and margin of error, i.e., p-value $<$ 0.05. This test suggested that a total of 192 failures would have been the sample that guarantees statistical representativeness of the population under these assumptions.

We opted to analyze half of the failures coming from violations from the symbolic oracles and the other half from metamorphic ones. These failures were randomly selected but we aimed at maximizing the VLA models, metamorphic relations and task types, as each may lead to different failure types (e.g., MR4 led to many failures not following the instruction). Afterwards, we developed a web-based application that permitted two experienced domain experts explain, in natural language, the failure type, i.e., the reason why the VLA failed. Each failure contained two videos of the simulation to the task, one for the source and the other for the follow-up test case. The task of annotating all the failures took around 1.5 hours to each of the two domain experts.

With this dataset, the first author of this paper developed a first version of the taxonomy of failures, by classifying the most recurring failure types. This taxonomy was subsequently validated and refined by another of the authors of the paper. Since LLMs have shown strong capabilities in reasoning over natural-language descriptions and identifying recurring patterns, three state-of-the-art LLMs, i.e., ChatGPT (GPT5.2 version), Gemini (Gemini 3 version) and Grok (Grok4 version) were used as assistants during the validation fo the taxonomy. The role of the LLMs in this phase was limited to suggesting refinements of the taxonomy. We acknowledge that LLMs can be prone to hallucination or errors. To mitigate these risks, we relied on multiple state-of-the-art LLMs and we conducted a thorough analysis of their outcomes allowing us to cross-check the proposed suggestions and reduce the influence of model-specific biases. All final decisions regarding the taxonomy were taken by the authors, with LLM feedback serving only as a supplementary tool to enhance the robustness, consistency and completeness of the resulting taxonomy.

\section{Analysis of the Results and Discussion}
\label{sec:results}
\subsection{$RQ_1$ -- Effectiveness}

\begin{figure*}[ht!]
    \centering
    \includegraphics[width=0.8\linewidth]{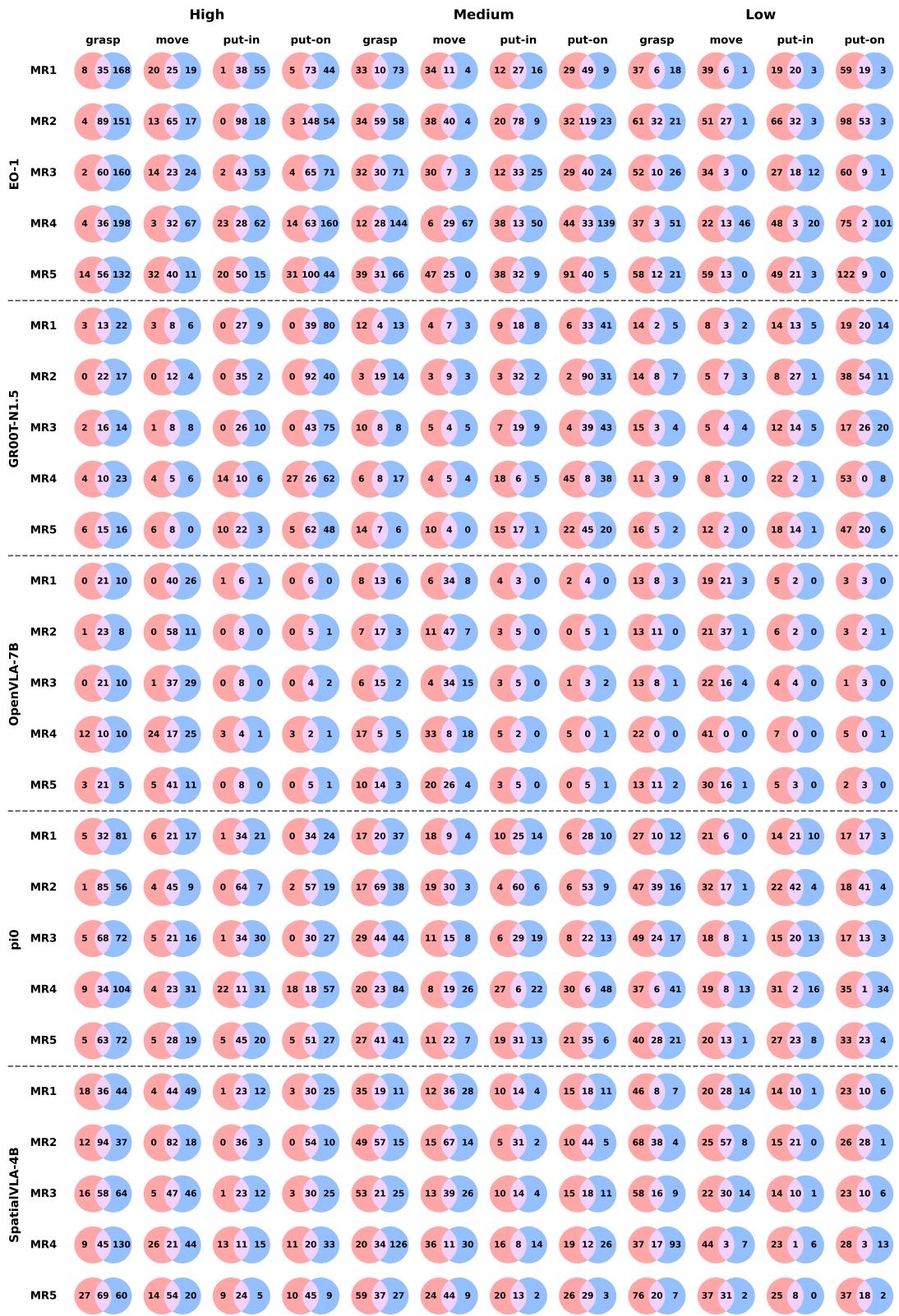}
    \caption{Venn diagram for each MR, model, threshold, and task combination showing the relationship between failures detected by the symbolic oracle and MR violations. The red circle represents failures detected only by the symbolic oracle, the blue circle represents MR violations not detected by the symbolic oracle, and the intersection (pink) represents MR violations as well as failures detected by the symbolic oracle}
    \label{fig:Venn}
\end{figure*}

Figure~\ref{fig:Venn} presents a Venn diagram for each model, task, MR, and threshold strictness level, summarizing failures~\footnote{Note that failures of symbolic oracles are genuine failures detected by these oracles, whereas failures related to MRs correspond to MR violations, which, while not necessarily goal-oriented failures, indicate deviations from the specified MR and therefore, potential faults.} detected by the symbolic oracle (in red), MT (in blue), and their overlap. At a high strictness threshold, MT showed a large number of MR violations, resulting in both a large intersection with failures detected by symbolic oracles and a high number of MR violations. In total, when considering a high threshold strictness level, for out of 7,822 failures, 640 were detected only by symbolic oracles, 3,527 by MRs, and 3,655 by both. This shows that MRs are effective at detecting failures, and both approaches are complementary, providing a more comprehensive evaluation of robot behavior. When considering a medium threshold of strictness, the total number of failures decreases to 6,286, distributed more evenly across detection categories: 1,776 failures are detected only by symbolic oracles, 1,991 only by MRs, and 2,519 by both. This indicates that, under a medium strictness level, MR violations are more closely aligned with correctness failures, while still contributing to behavioral inconsistency detection. 
Lastly, when considering a low threshold strictness level, the total number of detected failures further drops to 5,177. In this case, 2,877 failures are detected only by symbolic oracles, 8882 only by MRs, and 1,418 by both. This suggests that, with low strictness, MRs mainly reinforce symbolic failure detection rather than exposing additional behavioral deviations. 

\begin{custombox}{Answer to RQ1}
MT effectively detects failures in VLA-enabled robots, with its detection capability strongly influenced by the selected strictness threshold. Strict thresholds maximize the identification of additional failures beyond those captured by symbolic oracles, while less strict thresholds largely mirror oracle-based detection. Notably, the medium strictness threshold provides the best trade-off by aligning with failures detected by symbolic oracles while still exposing a significant number of additional failure cases.
\end{custombox}

\subsection{$RQ_2$ -- Comparison of MRs}

\begin{figure*}[ht]
    \centering
    \includegraphics[width=\linewidth]{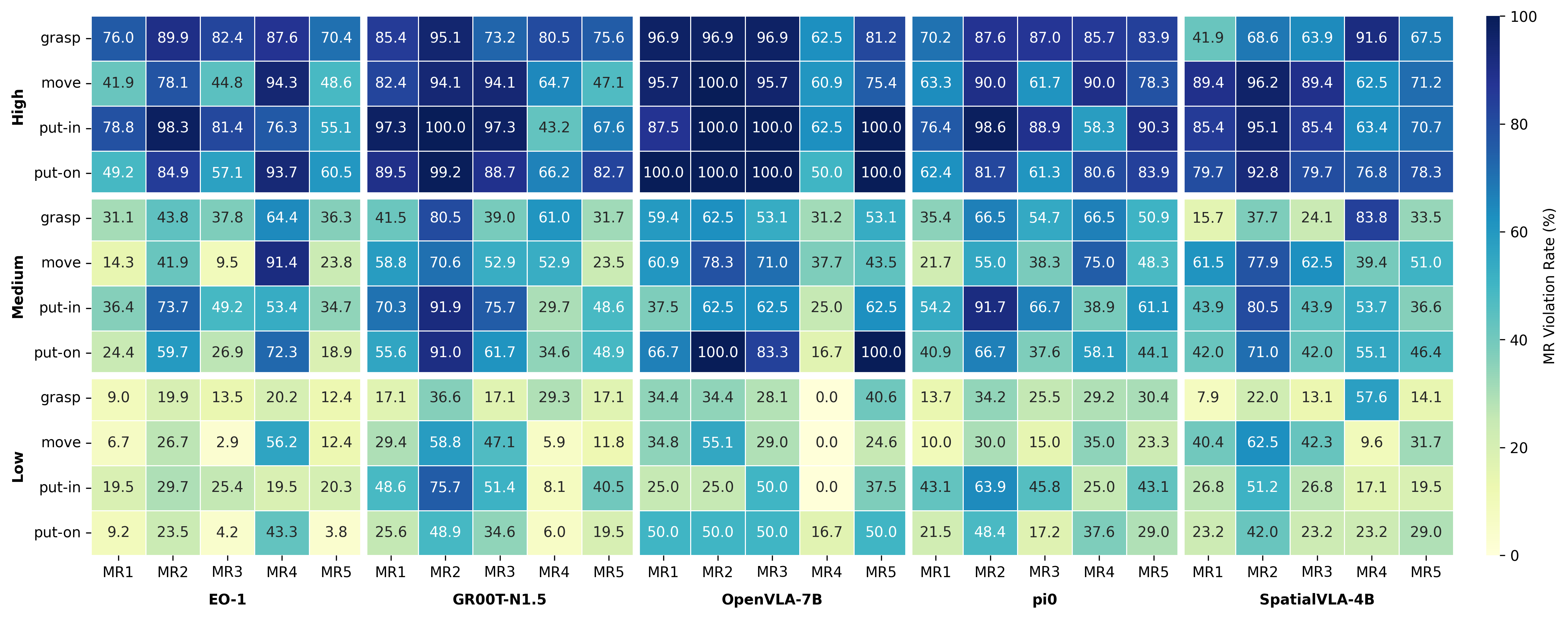}
    \caption{MR Violation Rate for each MR across three different strictness thresholds for each Model and task.}
    \label{fig:heatmap}
\end{figure*}

Figure~\ref{fig:heatmap} shows the MR violation rate for each MR across models, tasks, and strictness threshold levels. At the high threshold, all MRs tend to produce high violation rates, with MR2 showing the highest violation rate. Indeed, in most tasks and VLA models, MR2 showed the highest MR violation rate when the threshold strictness level was high. The only two exceptions were the Grasp task for the SpatialVLA-4B model and the move task for EO-1. While this strictness level might be too high for most applications, many robotic applications require extremely high precision. These results indicate that adding distracting objects in the workspace may degrade performance in applications requiring high precision.

At the medium strictness threshold, violation rates decrease across all MRs, but clear differences between relations remain. MR2 remains one of the most effective relations, particularly for tasks such as grasp and put-in. MR4 (Negation or Task Inversion) also shows relatively high violation rates under this setting, indicating that semantic transformations of the instruction often lead to inconsistent or unexpected behavioral changes. In contrast, MR1 (Synonym Substitution) and MR3 (Light Brightness Change) show more moderate violation rates, suggesting that while VLA models exhibit some robustness to lexical and visual perturbations, these transformations still reveal non-negligible behavioral inconsistencies in several models.

Under the Low threshold, violation rates are further reduced, emphasizing only the most significant deviations. MR2 again demonstrated the highest violation rate in most cases (14 out of 20), underscoring its robustness in detecting failures related to the performance effect of additional objects in the scene 
Exceptions include the move and put-on tasks for EO-1 (MR4 highest in both), the put-in task in OpenVLA-7B (MR3 highest), the move task in \pimodel~(MR4 highest), and the grasp task in SpatialVLA-4B (MR4 highest). 
At this level, MR3 and MR4 also emerge as notable in specific contexts, revealing weaknesses in environmental and instruction adaptation, respectively.



\begin{custombox}{Answer to RQ2}
Overall, environment-based transformations (MR2) and semantic transformations (MR4) are the most effective at exposing failures across models and tasks, while robustness-oriented relations, that is MR1 and MR3, are more sensitive to the chosen strictness threshold. This highlights the importance of combining multiple MRs to obtain a comprehensive assessment of VLA-enabled robotic behavior.
\end{custombox}

\subsection{$RQ_3$ -- Failure Type Characterization}

RQ3 investigates the types of failures revealed by different oracle types, namely symbolic oracles and metamorphic testing, and whether they expose complementary classes of failures. To answer this research question, we analyzed a statistically representative sample of failures drawn from both oracle types. We classified them using the taxonomy derived from expert annotations.
\begin{figure*}[ht]
    \centering
    \includegraphics[width=0.65\linewidth]{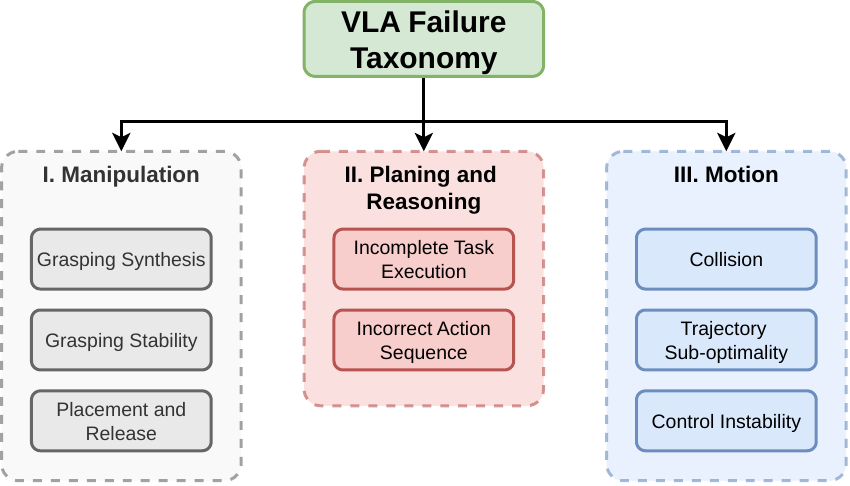}
    \caption{VLA Failure Taxonomy}
    \label{fig:Taxonomy}
\end{figure*}
Figure~\ref{fig:Taxonomy} presents the resulting taxonomy, which organizes VLA failures into three high-level categories: \emph{Manipulation}, \emph{Motion}, and \emph{Planning and Reasoning}. Each category is further refined into concrete failure types grounded in the observed robot behaviors during task execution. This taxonomy was derived from 192 annotated failures selected via Cochran’s sampling method and validated through expert agreement and LLM-assisted consolidation.

\noindent\textbf{\textit{Failures detected by symbolic oracles.}} Failures detected exclusively by symbolic oracles are predominantly concentrated in \emph{Planning and Reasoning} category. In particular, symbolic oracles frequently identify \emph{Incomplete Task Execution} and \emph{Incorrect Action Sequence} failures, where the robot does not reach the expected final symbolic goal state (e.g., the object is not lifted sufficiently high, not placed inside the container, or not released correctly). These failures correspond to violations of task-level correctness conditions encoded in the oracle definitions. 

However, symbolic oracles largely overlook failures that do not affect the final symbolic state. For instance, executions involving unstable grasps, minor collisions, or unnecessarily long and inefficient trajectories are often classified as successful by symbolic oracles, despite exhibiting degraded execution quality or safety-relevant issues.

\noindent\textbf{\textit{Failures detected by metamorphic testing.}} In contrast, MT reveals a broader diversity of failure types, with a strong presence in the \emph{Motion} and \emph{Manipulation} categories. Violations of MRs frequently expose \emph{Trajectory Sub-optimality}, \emph{Control Instability}, and \emph{Collision} failures, which manifest as oscillatory motions, abrupt corrections, or unintended contacts with objects or the environment. For example, under MR2 (Non-Interfering Object Addition), several VLA models exhibited unnecessary trajectory deviations caused by irrelevant objects, revealing sensitivity to non-task-related environmental changes. Similarly, MR5 (Target Object Relocation) revealed cases in which the robot failed to adapt its motion consistently and proportionally to the new target position, exposing deficiencies not only in high-level planning but also in motion safety and control. These failures are inherently behavioral and cannot be captured by binary symbolic goal conditions. 

The overlap between failures detected by symbolic oracles and MT mainly corresponds to severe failures affecting both task completion and execution behavior. However, a substantial portion of failures is detected exclusively by one oracle type. Symbolic oracles primarily capture \emph{what} the robot achieves, whereas MT captures \emph{how} the robot behaves.

Overall, the results show that MT exposes failure types that are largely invisible to symbolic oracles, particularly those related to execution quality, robustness, and safety. Conversely, symbolic oracles remain effective at identifying explicit goal violations. This complementarity confirms that relying on a single oracle type provides an incomplete view of VLA behavior.

\begin{custombox}{Answer to RQ3}
 Metamorphic testing reveals a more diverse set of failure types than symbolic oracles, particularly in the areas of motion quality and manipulation robustness. While symbolic oracles primarily detect task-level failures, MT uncovers execution-level and behavioral failures that are critical for assessing the reliability and safety of VLA-enabled robotic systems. These results demonstrate that combining symbolic oracles with MT is essential to achieve comprehensive failure coverage. 
\end{custombox}

\section{Threats to Validity}
\label{sec:threats}

In our evaluation, we sought to mitigate validity threats as thoroughly as possible. One \textbf{\textit{internal validity}} threat concerns potential heterogeneity in model selection and training. We mitigated this threat by selecting VLA models that were fine-tuned on the same dataset across all four tasks considered in the study, thereby reducing confounding effects due to differences in training data. Another internal validity threat relates to the choice of strictness thresholds used to determine metamorphic relation (MR) violations. To address this threat, we conducted a manual inspection of trajectory differences across thresholds, fine-tuning these thresholds manually with the help of domain experts. 

An \textit{\textbf{external validity}} threat concerns the generalizability of our findings beyond the specific models, tasks, and environments evaluated in this study. We mitigated this threat by evaluating our approach on five distinct VLA models and 1,864 source test cases spanning four different robotic tasks, thereby increasing the diversity of experimental conditions.

Finally, the construction of the failure taxonomy used to answer RQ3 may introduce a \textit{\textbf{conclusion validity}} threat, as it relies on qualitative judgment. To mitigate this threat, we involved two experienced domain experts in the failure analysis process and complemented their assessments with feedback from three state-of-the-art Large Language Models. Additionally, we developed a web-based tool to support consistent, systematic, and reproducible failure annotation and review.

\section{Related Work}
\label{sec:relatedwork}

Since its proposal, Metamorphic Testing (MT) has been widely applied to a broad range of systems, including compilers, machine-learning–based systems, and cyber-physical systems. Our work focuses on VLA–enabled robotic systems, which integrate multimodal AI techniques with physical robotic platforms. Within robotics, Laurent et al.~\cite{laurent2024metamorphic} applied MT to an autonomous delivery robot scheduler based on optimization algorithms. More generally, robotic systems can be viewed as Cyber-Physical Systems (CPS), for which Ayerdi et al.~\cite{ayerdi2022performance} proposed a performance-driven MT approach to assess system behavior beyond binary correctness. MT has also been successfully applied to other autonomous systems, including autonomous driving systems~\cite{zhou2022place,luu2024sequential,underwood2023metamorphic,luu2022sequential,zhang2023metamorphic}, drones~\cite{lindvall2017metamorphic}, and elevator control systems~\cite{ayerdi2020qos}. Our approach differs from existing robotics and CPS related MT techniques in several key aspects: (i) it targets end-to-end VLA-enabled robotic controllers, rather than isolated perception or planning components; (ii) it relies on trajectory-based MRs that capture qualitative and behavioral properties of task execution, rather than task-specific symbolic oracles; and (iii) it introduces generic, task-agnostic MRPs that enable systematic test generation across different robots, models, and tasks. 

In parallel with advances in VLA models, recent work has begun to address the problem of testing and evaluating VLA-enabled robotic systems. Most existing approaches rely on benchmark-based evaluations that assess task execution using symbolic or binary oracles. For example, VLATest~\cite{wang2025vlatest} proposes an automated framework for evaluating VLA models by generating task-specific symbolic oracles from predefined task and object templates. Other efforts, such as VLABench~\cite{zhang2025vlabench} and Nebula~\cite{peng2025nebula}, focus on large-scale benchmarking of VLA-enabled robots across a wide range of manipulation tasks and environments, reporting performance primarily in terms of task success rates. While these approaches provide valuable insights into the functional capabilities of VLA models, they largely focus on the correctness of the final task outcome. As a result, they offer limited support for assessing qualitative aspects of task execution, such as motion quality, manipulation robustness or perceptual grounding. 
Moreover, the reliance on symbolic or task-specific oracles makes these evaluations sensitive to linguistic variability and underspecified instructions. In contrast, our work targets the behavioral testing of VLA-enabled robotic systems, complementing existing evaluation methodologies by enabling the detection of execution-level and robustness-related failures that are not captured by traditional benchmark-based approaches.

\section{Conclusion and Future Work}
\label{sec:conclusion}

VLA models are emerging as a key control paradigm for generalist robots operating in unstructured environments. However, their evaluation remains challenging due to the oracle problem and the limited expressiveness of symbolic, goal-based correctness criteria. In this paper, we investigated the use of metamorphic testing as a complementary approach to assess the behavior of VLA-enabled robotic systems beyond final task success. We introduced two generic Metamorphic Relation Patterns and instantiated them into five concrete metamorphic relations. Unlike traditional symbolic oracles, these relations enable the assessment of execution-level properties. We empirically evaluated our approach on five state-of-the-art VLA models, two robotic platforms, and four representative manipulation tasks using more than 11,000 executions in simulation. Our findings suggest that symbolic oracles and metamorphic testing are complementary rather than competing approaches. Symbolic oracles remain well suited for assessing task-level correctness, while MT provides critical insight into how tasks are executed, capturing robustness, safety, and behavioral consistency. Relying on either oracle type in isolation yields an incomplete view of VLA behavior; combining both leads to substantially broader failure coverage and a more realistic assessment of system reliability.

In the future, we plan to extend the paper from different perspectives: First, we want to extend the work and involve humanoid robots. Second, we would like to include further metamorphic relations, and employ metamorphic relation composition \cite{qiu2020theoretical} to reduce the testing cost. Lastly, we intend to investigate automated and adaptive calibration of metamorphic relation thresholds to reduce manual tuning and improve generalizability across robots, tasks, and environments. 

\section*{Replication Package}

For the sake of replicability, we provide a full replication package~\cite{MT_replication_package}. In addition, the full implementation of the approach is available on the project webpage: \textcolor{blue}{\url{https://pablovalle.github.io/MT_of_VLAs_web/}}

\section*{Acknowledgments} 
Pablo Valle and Aitor Arrieta are part of the Software and Systems Engineering research group of Mondragon Unibertsitatea (IT1519-22), supported by the Department of Education, Universities and Research of the Basque Country. Pablo Valle is supported by the Pre-doctoral Program for the Formation of Non-Doctoral Research Staff of the Education Department of the Basque Government (Grant n. PRE\_2025\_2\_0252). Shaukat Ali is supported by the Co-tester project (No. 314544) funded by the Research Council of Norway. This work has been partially funded by the Spanish Ministry of Science, Innovation and Universities (project PID2023-152979OA-I00), funded by (MCIU /AEI /10.13039/501100011033/FEDER, UE). This paper is also part of the project PID2024-156482NB-I00, funded by MICIU/AEI/10.13039/501100011033 and by the ESF+.

\ifCLASSOPTIONcaptionsoff
  \newpage
\fi

\bibliographystyle{ieeetr}
\bibliography{metamorphic}


\end{document}